\documentclass[letterpaper,10pt,conference]{ieeeconf}
\IEEEoverridecommandlockouts
\let\labelindent\relax
\makeatletter
\let\FELTIEEEcaption\@makecaption
\makeatother
\usepackage{amssymb}
\usepackage{graphicx}
\usepackage{booktabs}
\usepackage{makecell}
\usepackage{verbatim}
\usepackage{enumitem}
\usepackage{pifont}
\usepackage{xspace}
\usepackage{colortbl}
\usepackage{xcolor}
\usepackage{ragged2e}
\usepackage{amsmath} 
\usepackage{subcaption}
\usepackage{multirow}
\definecolor{finalcol}{gray}{0.90}
\newcommand{\method}{FELT\xspace}

\newcommand{\ba}{\mathbf{a}}

\newcommand{\bz}{\mathbf{z}}
\newcommand{\cmark}{\ding{51}} 
\newcommand{\xmark}{\ding{55}}

\usepackage{cite}
\usepackage[bookmarks=false,hidelinks]{hyperref}
\makeatletter
\def\bstctlcite{\@ifnextchar[{\@bstctlcite}{\@bstctlcite[@auxout]}}
\def\@bstctlcite[#1]#2{\@bsphack
  \@for\@citeb:=#2\do{%
    \edef\@citeb{\expandafter\@firstofone\@citeb}%
    \if@filesw\immediate\write\csname #1\endcsname{\string\citation{\@citeb}}\fi}%
  \@esphack}
\makeatother
\hypersetup{pdfauthor={Zinan Li, Yiyang Ling, Yuming Gu, Binghao Huang, Chenhao Liang, Sharfin Islam, Hisham Bedri, John Chirikjian, Yunzhu Li, Stefanos Nikolaidis, Daniel Seita},pdftitle={FELT: Generating Tactile Signals from Vision for Visuo-Tactile Manipulation}}
\title{\LARGE\bf FELT: Generating Tactile Signals from Vision\\for Visuo-Tactile Manipulation}
\author{%
\authorblockN{\normalsize
Zinan Li\textsuperscript{1,*}, Yiyang Ling\textsuperscript{1,*}, Yuming Gu\textsuperscript{1,\textdaggerdbl}, Binghao Huang\textsuperscript{2}, Chenhao Liang\textsuperscript{1}\\
Sharfin Islam\textsuperscript{3}, Hisham Bedri\textsuperscript{3}, John Chirikjian\textsuperscript{3}, Yunzhu Li\textsuperscript{2}, Stefanos Nikolaidis\textsuperscript{1}, Daniel Seita\textsuperscript{1}}
\authorblockA{\small
\textsuperscript{1}Viterbi School of Engineering, University of Southern California\quad \textsuperscript{2}Columbia University\quad \textsuperscript{3}Starpilot\\
\textsuperscript{*}Equal contribution\quad \textsuperscript{\textdaggerdbl}Project lead\quad Correspondence: \texttt{zinanl@usc.edu}, \texttt{seita@usc.edu}}}
\begin{document}
\makeatletter
\long\def\@makecaption#1#2{%
  \ifx\@captype\@IEEEtablestring
    \vskip\abovecaptionskip
    {\footnotesize\normalfont
     \leftskip=0pt\rightskip=0pt\parfillskip=0pt plus 1fil
     \noindent #1:~#2\par}%
  \else
    \FELTIEEEcaption{#1}{#2}%
  \fi
}
\makeatother
\bstctlcite{FELT:BSTcontrol}
\IEEEaftertitletext{%
\begin{minipage}{\textwidth}
\centering
\captionsetup{type=figure}
\includegraphics[width=1.0\linewidth]{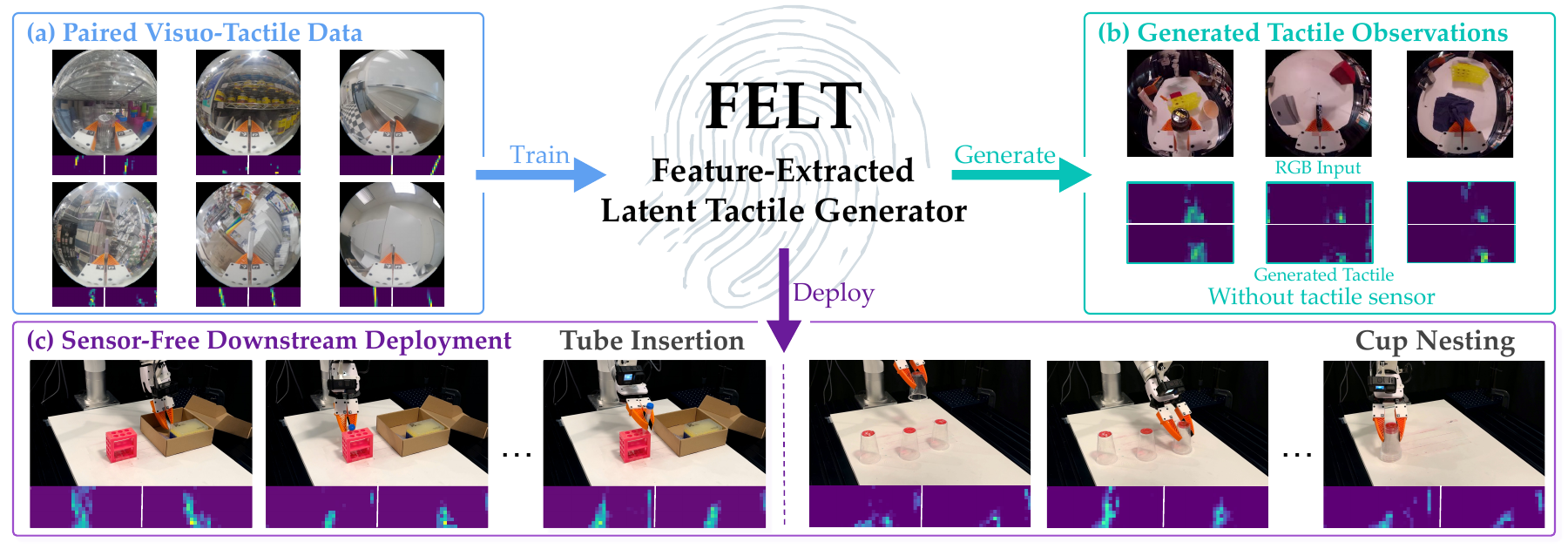}
\caption{
  \method uses existing paired visuo-tactile data~\cite{zhu2025touchwild} (a) to learn spatial contact representations from RGB observations, with examples of generated tactile images shown in (b). For downstream manipulation (c), the image and feature interfaces provide generated tactile images or contact-aware features to policies, enabling deployment without real tactile inputs.
}
\label{fig:teaser}
\end{minipage}\endgraf
\vspace{\baselineskip}
}
\maketitle
\thispagestyle{empty}
\pagestyle{empty}
\begin{abstract}
Tactile sensing can improve performance in contact-rich robot manipulation. However, learning policies that use real tactile feedback typically requires demonstrations collected with tactile-equipped hardware.
We present \textbf{\underline{F}}eature-\textbf{\underline{E}}xtracted \textbf{\underline{L}}atent \textbf{\underline{T}}actile (\method), a framework that uses existing paired visuo-tactile data to learn contact-aware representations from RGB observations. \method provides these representations to downstream policies through two interfaces: generated tactile images and contact-aware features.
\method combines a frozen visual encoder with separate query decoders for the left and right finger, reflecting the spatial organization of dual-finger tactile arrays. The two branches exchange information through gated connections, and a convolutional readout produces spatial contact representations for the downstream interfaces.
For downstream manipulation, the image interface decodes these spatial representations into tactile images, allowing policies trained with either real or generated tactile observations to use generated images at deployment. Alternatively, the feature interface feeds these representations directly to the policy during both training and deployment, without real tactile observations.
Experiments on four contact-rich real-robot tasks show that both interfaces achieve higher success rates than vision-only policies. Directly incorporating learned contact-aware features into the policy improves success by 15--22.5 percentage points across the four tasks.
Website: \nolinkurl{https://felt-tactile.github.io/}.
\end{abstract}

\section{Introduction}

Robotic manipulation has seen substantial progress~\cite{real_world_robot_FMs_2024,FMs_robotics_2023_survey_1,FMs_robotics_2023_survey_2}, but real-world deployment in unstructured environments remains limited by \textbf{the high cost of collecting diverse robotic data}. Current approaches for large-scale imitation learning rely on expensive teleoperation data~\cite{open_x_embodiment_rt_x_2023} with specialized hardware and expert operators, and typically capture only visual observations.
Yet, recent work has shown that vision and tactile sensing provide essential complementary information: vision offers global context, while tactile feedback delivers local, high-resolution contact signals~\cite{huang2024_3DViTac,guzey2024touch,RobotSynesthesia2024}.
However, tactile data is particularly challenging to collect, as it requires specialized, fragile sensors and careful calibration~\cite{zorin2026tacobenchmarkingtactile}.
Visual observations provide cues about where and how the robot contacts objects, but similar RGB data can correspond to different tactile readings.
Existing paired visuo-tactile data~\cite{zhu2025touchwild} therefore offers a way to use tactile supervision to learn contact-aware representations from these signatures.
After this pretraining, the learned model supports data augmentation by adding predicted tactile images or contact-aware features to vision-only data, without requiring new tactile measurements for downstream manipulation.

We present \textbf{\underline{F}}eature-\textbf{\underline{E}}xtracted \textbf{\underline{L}}atent \textbf{\underline{T}}actile (\method), a framework that learns spatial, per-finger contact representations from RGB observations and provides two interfaces for downstream manipulation: generated tactile images and contact-aware features.
\method learns from existing paired visuo-tactile demonstrations~\cite{zhu2025touchwild} and structures its contact representations around the spatial layout of tactile arrays mounted on the two fingers of a parallel-jaw gripper, a sensing configuration widely used in many visuo-tactile manipulation systems~\cite{huang2024_3DViTac}.
The tactile sensors on the left and right fingers capture distinct but related contact patterns during manipulation.
To capture this structure, \method combines a frozen visual encoder with separate query decoders for each finger, connected through gated information exchange. A  convolutional readout then transforms the decoded features into spatial contact representations for the two downstream interfaces.
The image interface allows policies trained with either real or generated tactile observations to use generated tactile images at deployment. The feature interface provides contact-aware features directly to policies during training and deployment, bypassing image generation and re-encoding without requiring real tactile observations.

We evaluate both interfaces with Diffusion Policies~\cite{chi2025diffusion,zhu2025touchwild} on four contact-rich real-robot tasks, comparing against vision-only policies and policies using real tactile observations. Both generated tactile images and contact-aware features improve policy success rates over vision-only policies across all four tasks. See Figure~\ref{fig:teaser} for an overview of the setup and downstream visuo-tactile tasks.

Our contributions are: 
\textbf{(1)} \method, a framework that generates per-finger tactile images and contact-aware features from RGB observations by modeling each finger's contact patterns separately while sharing information between the two fingers;
\textbf{(2)} two downstream interfaces: generated tactile images for deploying policies trained with either real or generated tactile observations, and contact-aware features for policy training and deployment without real tactile inputs; and
\textbf{(3)} real robot experiments across four contact-rich manipulation tasks demonstrating improved policy success over vision-only baselines with both interfaces, supported by tactile prediction evaluations, architectural ablations, and pretraining data analyses.

\section{Related Work}
\label{sec:related_work}

\noindent \textbf{Robotic Tactile Sensors.}
Recent years have seen great advances in tactile sensors for robotic manipulation. These range from smaller magnetic~\cite{bhirangi2021reskin,bhirangi2025anyskin,pattabiraman2025eflesh}, or gel-based~\cite{donlon2018gelslim,sipos2024gelslim40focusingtouch} to fingertip sensors~\cite{patel2021diggerfingergelsighttactile,lambeta2024digitizingtouch}, as well as handheld devices~\cite{zhu2025touchwild,cheng2026tacumi}. 
These tactile sensors can be utilized for delicate, fine-grained tasks such as separating layers of fabric~\cite{tirumala2022} and picking up papers~\cite{lin2025pptacpaperpicking}. 
Recently, researchers have also developed full robotic hands~\cite{romero2024eyesight,zhou2025motifhandrobotichand} and skins~\cite{si2023robotsweater} that provide tactile feedback across larger contact areas. 
While there have been great efforts at expanding open-source tactile sensors~\cite{huang2024_3DViTac,huang2026flexitac}, benchmarking tactile sensors~\cite{zorin2026tacobenchmarkingtactile}, and enabling cross-sensor generalization~\cite{rodriguez2025crosssensor}, collecting large-scale tactile datasets still requires substantial manual effort. 
Our work complements these efforts: instead of designing a new sensor, we reuse existing paired visuo-tactile data to augment vision-based policies with learned contact representations.

\noindent \textbf{Tactile and Visuo-Tactile Manipulation.}
Prior work on \underline{\smash{visuo-tactile manipulation}} shows that combining vision and touch improves performance over using either modality alone~\cite{huang2025tactilevla,bi2025vlatouch,guzey2023dexteritytouch}, due to their complementary strengths~\cite{zhao2025touchbeginsvisionends}. 
Visuo-tactile policies use images~\cite{lee2019visiontouch} or point clouds~\cite{huang2024_3DViTac} for global scene understanding and tactile signals for local contact geometry, which is critical for contact-rich tasks such as peg insertion~\cite{chen2022visuotactiletransformers,guzey2024touch}, table wiping~\cite{jiang2025gelfusion}, or in-hand rotation~\cite{RobotSynesthesia2024,qi2023general,yin2023rotating}. 
Other works have studied visuo-tactile policies for bimanual manipulation using multifingered hands~\cite{lin2025learning}. 
However, data collection remains a bottleneck, where existing large-scale datasets~\cite{open_x_embodiment_rt_x_2023,khazatsky2024droid} lack tactile information.
Our work is complementary to policy learning efforts: trained on existing paired visuo-tactile data, \method provides tactile images compatible with visuo-tactile policies and features for downstream training and deployment without further tactile data collection.

\noindent \textbf{Data Augmentation and Tactile Supervision.}
In parallel, \underline{\smash{data augmentation}} has become widely utilized in robot learning to synthetically expand training data at a lower cost compared to teleoperation~\cite{ji2026oxeaugelargescale}. Existing methods expand visual data by randomizing backgrounds or objects~\cite{bharadhwaj2024roboagent,chen2024semanticallycontrollable}, synthesizing viewpoints~\cite{zhou2023nerfpalmhand,zhang2024diffusionmeetsdagger,liu2025DCODA,tian2024vista}, or learned world models~\cite{guo2025ctrlworldcontrollablegenerativeworld,jang2025dreamgenunlockinggeneralizationrobot,chen2026craft}. 
Visuo-tactile world models additionally show that tactile observations can improve imagined rollouts and planning~\cite{higuera2026visuotactileworldmodels}.  
More recent work uses action-conditioned visuo-tactile rollouts for data augmentation~\cite{huang2026vitacworld}, or combines future tactile prediction with real-time tactile feedback for action correction~\cite{ma2026tacpac}.
We use tactile supervision to augment existing RGB observations with contact representations for downstream policy learning.
ControlTac~\cite{luo2025controltac} generates tactile images by warping a single fixed reference tactile observation, while ViTacGen~\cite{wu2025vitacgenroboticpushingvisiontotouch} uses vision-to-touch generation for robotic pushing. 
Other work studies cross-sensor tactile generation, translating touch signals between heterogeneous tactile sensors~\cite{rodriguez2025crosssensor}. 
As concurrent work, TacImag~\cite{zhang2026tacimag} predicts tactile observations from vision and proprioception to guide manipulation policies without tactile sensors at deployment, while TacGen~\cite{ye2026tacgen} combines visuo-tactile contrastive alignment with vision-to-touch latent generation.
Our aim is to support multiple manipulation tasks with spatial, per-finger contact representations learned from RGB, without requiring real tactile inputs at deployment.

\begin{figure*}[t]
\centering
\includegraphics[width=\linewidth]{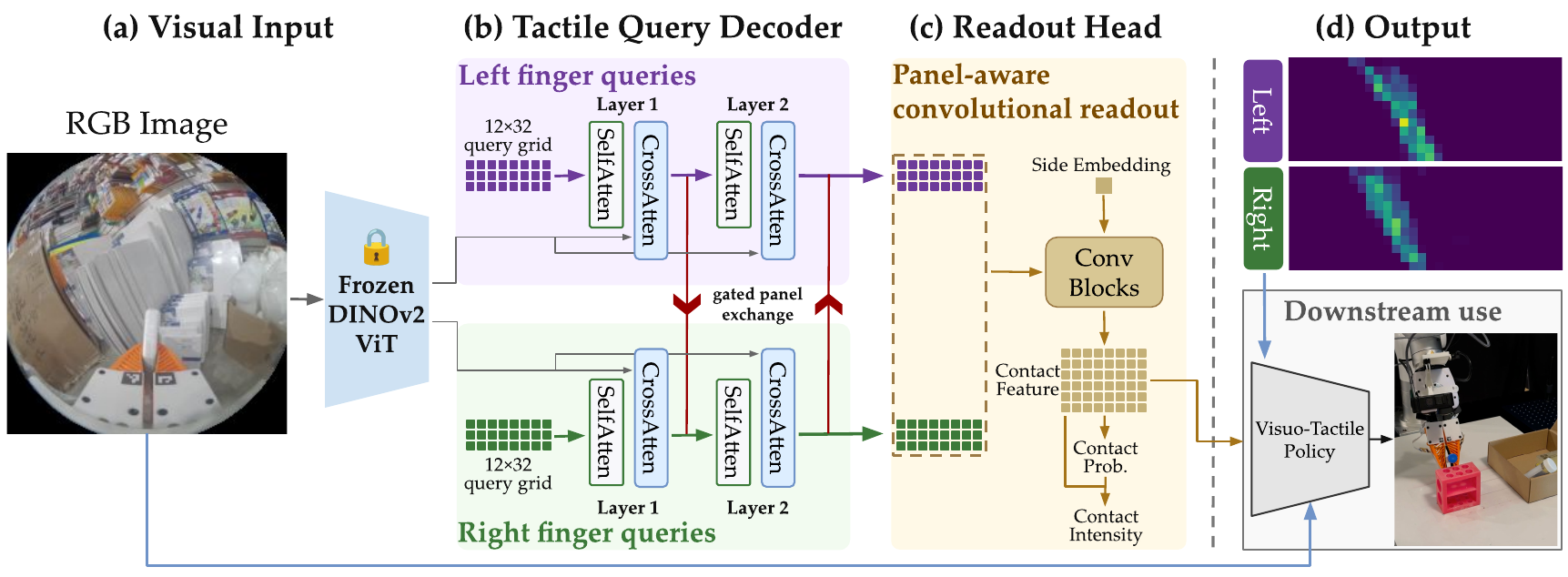}
\caption{
Overview of \method's spatial, per-finger contact representation framework (see Section~\ref{sec:method}).
(a) Given a fisheye RGB observation~\cite{chi2024umi}, \method extracts visual features using a frozen, pretrained DINOv2 ViT~\cite{oquab2024dinov2learningrobustvisual,dosovitskiy2021imageworth16x16words}.
(b) Separate decoders for the left and right finger panels use learnable, position-aware queries aligned with each panel's spatial grid. Each branch attends to the visual features through cross-attention, while gated cross-panel connections enable information exchange between the two branches.
(c) A panel-aware convolutional readout transforms the decoded queries into spatial contact features, with learned side embeddings to distinguish the left and right panels. Prediction heads map these features to per-cell contact probabilities and conditional contact intensities.
(d) \method provides two interfaces for downstream manipulation: generated tactile images and contact-aware features.
} %
\label{fig:pipeline}
\vspace{-10pt}
\end{figure*}

\section{Problem Statement and Assumptions}
\label{sec:problem_statement}

To learn contact representations for downstream manipulation, we begin with a dataset of $N$ paired visuo-tactile demonstrations $\mathcal{D} = \{\tau_i\}_{i=1}^N$, where each trajectory $\tau$ consists of a sequence of visuo-tactile-action tuples $(I_t^\text{vis}, I_t^\text{tac}, \ba_t)$.
Here, $I_t^\text{vis} \in \mathbb{R}^{H\times W \times 3}$ is the RGB image, and $I_t^\text{tac} \in \mathbb{R}^{H' \times W'\times 2}$ contains a unitless contact-intensity map for each of the two finger sensors~\cite{huang2024_3DViTac}, which we refer to as a sensor panel. The two panels correspond to the left and right fingers, with $H'{=}12$ and $W'{=}32$ in our setup. $\ba_t$ denotes the robot action at time $t$.
Our goal is to learn a framework from $\mathcal{D}$ that provides generated tactile images and contact-aware features from RGB observations for downstream policy learning and deployment.
Formally, we learn a generator $G_\phi$ with parameters $\phi$ that predicts tactile images $\widetilde{I^\text{tac}_t} = G_\phi(I_t^\text{vis})$. We denote its intermediate per-finger spatial features by $(\bz_t^L,\bz_t^R) = G_\phi^\text{feat}(I_t^\text{vis})$, which provide an alternative policy interface without explicit tactile-image generation.

\section{Method}
\label{sec:method}

\subsection{Attention-Based Tactile Generation Network}
\label{ssec:part1}

To implement $G_\phi$ (see Figure~\ref{fig:pipeline}), we introduce an attention-based~\cite{vaswani2017attentionneed} cross-modal framework that predicts per-finger contact probabilities and contact intensities.

\noindent \textbf{\method Architecture.} Given RGB observations $I_t^\text{vis}$ from a fisheye camera setup~\cite{chi2024umi,zhu2025touchwild}, we center-crop the images and resize them to $224 \times 224$. $G_\phi$ extracts visual tokens using a frozen pretrained DINOv2-B/14~\cite{oquab2024dinov2learningrobustvisual} Vision Transformer~\cite{dosovitskiy2021imageworth16x16words}, averaging features from multiple network depths to capture both low-level texture and high-level semantic information.
Since the left and right finger panels are separate surfaces that experience distinct contact patterns, we assign each panel its own set of 12$\times$32 learnable query embeddings matching the physical sensor grid, augmented with 2D positional embeddings via row/column factorization. Each branch uses an independent two-layer transformer decoder with a hidden dimension of 768. Within each layer, queries first attend to one another through self-attention and then attend to the visual tokens through cross-attention. After each decoder layer, a gated cross-panel exchange module allows the two branches to share information, enabling the model to reason about coordinated contact.
A shared panel-aware \emph{convolutional readout head} reshapes each branch's decoder queries into 2D grids and processes them through residual depthwise-separable convolution blocks with learned side embeddings that distinguish left from right. A contact head predicts per-cell contact probabilities $\hat{c}_t$ from the convolutional features $\bz_t^L,\bz_t^R$. These probabilities are concatenated with the features and passed to an intensity head to produce the final nonnegative tactile intensity map $\widetilde{I_t^\text{tac}}$.

\noindent \textbf{Training.} We derive the per-cell binary contact labels $c_t$ by thresholding $I_t^\text{tac}$ at 0.01. We train $G_\phi$ on paired demonstrations $\mathcal{D}$ by minimizing a composite loss for contact detection and contact intensity estimation:
\begin{equation}
\label{eq:loss}
\begin{aligned}
\mathcal{L}(\phi)
=\mathbb{E}_{(I_t^\text{vis},I_t^\text{tac})\sim\mathcal{D}}\Big[\,
    &\mathcal{L}_\text{contact}(\hat{c}_t,c_t)\\
    &+\mathcal{L}_\text{intensity}(\widetilde{I_t^\text{tac}},I_t^\text{tac})\,\Big].
\end{aligned}
\end{equation}
The contact objective combines focal loss~\cite{lin2017focal}, soft Dice loss~\cite{milletari2016vnet} with weight 0.5, and an $L_2$ penalty on the contact logits with weight 0.01. The intensity objective combines value-weighted Huber loss~\cite{huber1964robust} with background suppression and shape regularization: $\mathcal{L}_\text{intensity}=\mathcal{L}_\text{Huber}+\lambda_1\mathcal{L}_\text{bg}+\lambda_2\mathcal{L}_\text{shape}$, where $\lambda_1=\lambda_2=0.05$ in all experiments. The value-weighted Huber term emphasizes high-intensity contact cells. The background suppression term $\mathcal{L}_\text{bg}$ penalizes predicted intensity outside the dilated ground-truth contact region. The shape term $\mathcal{L}_\text{shape}$ measures the $L_1$ distance between predicted and ground-truth intensity maps, each normalized by its total intensity with a small constant for numerical stability.

\subsection{Deploying the Tactile Generator}
\label{ssec:augment_data}

The generator provides two downstream policy interfaces: generated tactile images and contact-aware features.

\noindent \textbf{Image interface.} Given RGB observations $I_t^\text{vis}$, we generate tactile images $\widetilde{I_t^\text{tac}}=G_\phi(I_t^\text{vis})$. The policy's tactile encoder maps either real tactile images $I_t^\text{tac}$ or generated images $\widetilde{I_t^\text{tac}}$ to a 768-dimensional tactile token. Generated tactile images can replace sensor readings at deployment for a policy trained with real tactile observations, or augment vision-only demonstrations for policy training with generated images used at both training and deployment. The pretrained tactile generator remains frozen during downstream policy training and deployment.

\noindent \textbf{Feature interface.} Given RGB observations $I_t^\text{vis}$, we extract per-panel spatial features $(\bz_t^L,\bz_t^R)=G_\phi^\text{feat}(I_t^\text{vis})$ before the final prediction heads, where $\bz_t^L,\bz_t^R\in\mathbb{R}^{12\times32\times256}$. Each map is spatially mean-pooled; the pooled vectors are concatenated and linearly projected into a 768-dimensional tactile token. This token replaces the tactile encoder's output, bypassing image prediction and re-encoding. During policy training, the DINOv2 encoder remains frozen while the tactile decoder is fine-tuned jointly with each task-specific policy. This interface requires no real tactile input during downstream training or deployment.

\section{Experiments}
\label{sec:experiments}

\subsection{Hardware and Datasets}
\label{ssec:datasets}

\noindent \textbf{Hardware.} Experiments are conducted on an xArm7 with a UMI-style gripper~\cite{chi2024umi} instrumented with FlexiTac, flexible piezoresistive tactile sensor pads that provide per-finger $12 \times 32$ contact-intensity maps~\cite{huang2024_3DViTac,zhu2025touchwild,huang2026flexitac}. RGB observations are captured from a wrist-mounted GoPro Hero 9 with an Ultra Wide Lens Mod, following~\cite{chi2024umi}. RGB, robot states, and tactile readings are aligned to a common RGB timeline after correcting capture and sensor timestamp offsets; episodes with an RGB--tactile matching gap exceeding 100\,ms are excluded. The xArm setup uses different FlexiTac units from those used to collect the generator pretraining data, with per-session recalibration to correct offsets and drift~\cite{zhu2025touchwild}.

\noindent \textbf{Generator Datasets.}
We train the tactile generator on the paired visuo-tactile dataset from Zhu~et~al.~\cite{zhu2025touchwild}, which contains over 2,700 handheld UMI-style demonstrations and 2.6M RGB-tactile pairs. We use an 80/20 train/evaluation split by demonstration. The policy encoder is also pretrained on this dataset~\cite{zhu2025touchwild}. To test generator quality beyond the training distribution, we additionally collect an xArm test set of 30 GELLO~\cite{wu2024gello} demonstrations with ${\sim}$72K visuo-tactile frames, using objects and task scenarios not seen during generator training.

\noindent \textbf{Policy Dataset.} For downstream policy learning, we collect 60 GELLO demonstrations per task on the xArm, totaling 240 demonstrations and ${\sim}$520K synchronized RGB, tactile, and action frames across four tasks. These demonstrations are collected independently from both the generator training data and the xArm generator test set.

\subsection{Visuo-Tactile Manipulation Tasks}
\label{ssec:tasks}

Across four evaluation tasks, we initialize \method from the same pretrained generator checkpoint and train separate task-specific visuo-tactile Diffusion Policies~\cite{zhu2025touchwild,chi2025diffusion} on the demonstrations from Section~\ref{ssec:datasets}. The tasks are:
\begin{enumerate}[noitemsep,leftmargin=*]
    \item \textbf{Tube Insertion~\cite{zhu2025touchwild}.} The robot grasps a tilted test tube, reorients it using the environment, and inserts it into a receptacle. Success requires stable insertion.
    \item \textbf{Cup Nesting.} The robot sequentially picks up four upside-down cups and nests them into a single stack. Success requires all four cups to be fully nested.
    \item \textbf{Eraser Wiping.} The robot grasps an eraser from a cup and removes two marker lines from the surface. Success requires both lines to be completely erased within five wiping passes.
    \item \textbf{Triangle Peg Insertion.} The robot grasps a triangular peg with a randomized initial orientation, reorients it, and inserts it into a matching triangular hole. Success requires stable insertion.
\end{enumerate}

\subsection{Experiment Protocol and Evaluation of \method}
\label{ssec:exp_evals}

We evaluate tactile prediction quality, downstream policy success, architectural ablations, and frozen representations. %

We train the tactile generator for 40 epochs using AdamW with a learning rate of $2\times10^{-4}$ and batch size 128, selecting the checkpoint with the lowest validation tactile-image MAE.

\noindent \textbf{Tactile Prediction Quality.}
We evaluate generated tactile maps against synchronized sensor readings on the xArm test set. Because tactile maps are sparse, SSIM~\cite{wang2004image} and PSNR can be dominated by non-contact background regions and may underrepresent contact-region errors. We therefore report four complementary metrics. (1) \underline{\smash{Learned Perceptual Image Patch Similarity (LPIPS)}}~\cite{zhang2018unreasonable} measures perceptual discrepancy; single-channel maps are replicated across three channels and bilinearly upsampled before evaluation. (2) \underline{\smash{Energy Ratio}} is the mean per-frame ratio of predicted to ground-truth total contact intensity over frames with nontrivial ground-truth intensity, with each ratio capped at 10; values closer to 1 are better. (3) \underline{\smash{Contact Accuracy}} measures the fraction of finger panels whose predicted binary contact state matches the ground truth; a panel is classified as in contact when at least 12 cells exceed an intensity of 0.2. (4) \underline{\smash{Panel Asymmetry Error}} measures the absolute difference between predicted and ground-truth log ratios of left-to-right mean panel intensity, averaged over contact frames; a small constant is added to panel intensities for numerical stability.

We compare \method against three tactile prediction baselines, each receiving only RGB observations at inference. \underline{\smash{Nearest Neighbor}} retrieves the training frame with the highest cosine similarity between mean-pooled frozen DINOv2 features and returns its paired tactile image. \underline{\smash{Masked Autoencoder (MAE)}}~\cite{he2022masked} uses the visuo-tactile MAE architecture from Zhu~et~al.~\cite{zhu2025touchwild}, with all tactile tokens masked during both training and evaluation so that reconstruction relies on RGB alone. We train \underline{\smash{ViTacGen}}~\cite{wu2025vitacgenroboticpushingvisiontotouch} on our dataset with input and output shapes adjusted to match our data. We use the same checkpoint-selection criterion as \method and evaluate it on the xArm test set.

\begin{figure*}[t]
\centering
\includegraphics[width=\linewidth]{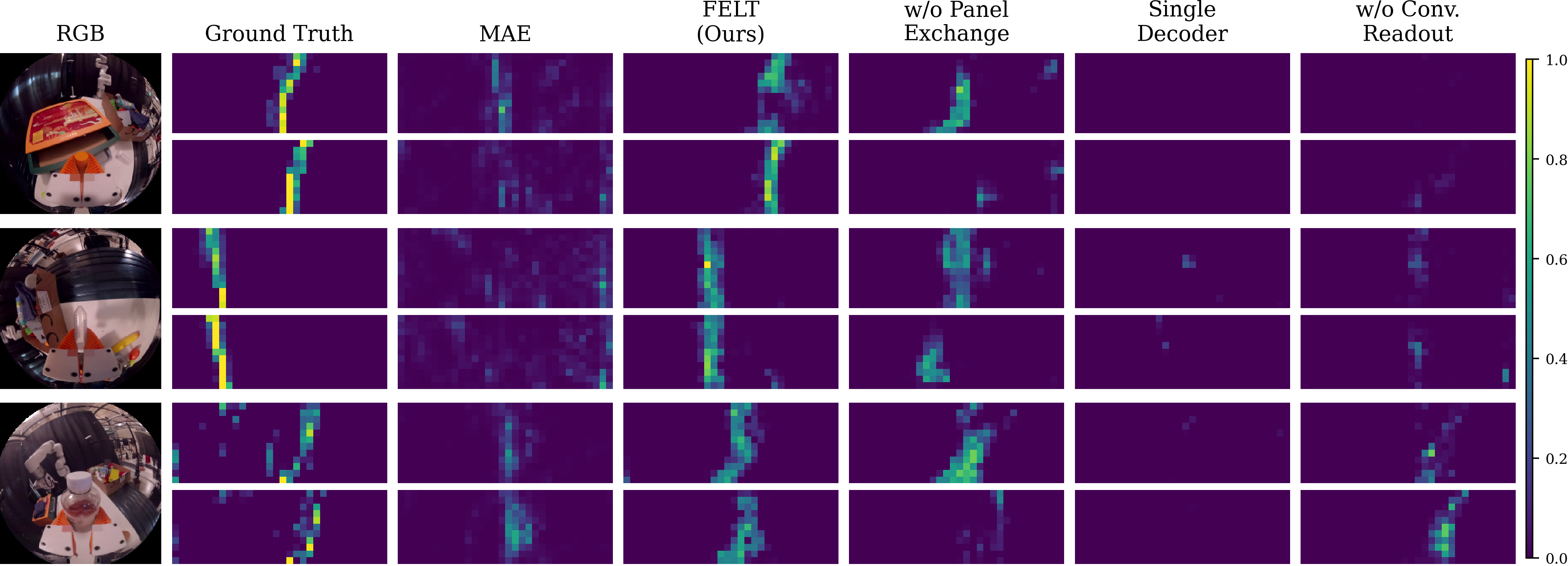}
\caption{
  \textbf{Qualitative tactile prediction and ablation results on the xArm test set.}
  Each row shows, from left to right, the RGB observation, ground-truth tactile image, and predictions from MAE, the full \method model, and its ablated variants. Tactile images display the left and right finger panels stacked vertically. The single-decoder and no-convolutional-readout variants underestimate contact intensity, while removing panel exchange distorts the relative intensity between the left and right panels.
}
\label{fig:xarm_ablation}
\vspace{-10pt}
\end{figure*}

\noindent \textbf{Downstream Policy Learning.}
We evaluate \method using task-specific Diffusion Policies~\cite{chi2025diffusion} built on the pretrained visuo-tactile encoder from Zhu~et~al.~\cite{zhu2025touchwild}. All configurations share the same RGB observations, proprioceptive inputs, CLIP visual encoder~\cite{radford2021clip}, and diffusion policy backbone, and are trained for 60 epochs under the same optimization protocol. We compare six configurations that differ in the additional representation supplied to the policy:
(1) \underline{\smash{Vision Only}}: no additional representation during policy training or deployment;
(2) \underline{\smash{DINOv2 Features}}: mean-pooled, linearly projected frozen DINOv2 features during both policy training and deployment;
(3) \underline{\smash{Real Tactile}}: real tactile observations during both policy training and deployment;
(4) \underline{\smash{\method Images (Real$\rightarrow$Predicted)}}: real tactile observations during policy training, replaced by \method-generated images at deployment;
(5) \underline{\smash{\method Images (Predicted$\rightarrow$Predicted)}}: \method-generated tactile images during both policy training and deployment; and
(6) \underline{\smash{\method Features}}: intermediate generator features during both policy training and deployment, without real tactile observations at either downstream stage.

Each of the six configurations is evaluated over 40 trials per task.
Trials are interleaved across configurations with randomized execution order to reduce ordering effects and operator reset bias~\cite{kressgazit2024robotlearning}. Each trial starts from a fixed robot pose with a randomized object configuration and lasts at most 120\,s. Policies execute actions at 10\,Hz. Policies use two-frame observations and predict 16-step action chunks, executing the first six actions before replanning. We report final task success rates in Table~\ref{tab:policy_main}.

\noindent \textbf{Ablations.}
We evaluate three architectural variants of the tactile generator while keeping the visual encoder, training losses, and training schedule unchanged:
(1) \underline{\smash{\method w/o panel exchange}}, which removes gated cross-panel attention while retaining the separate per-finger decoders and convolutional readout;
(2) \underline{\smash{\method single-decoder variant}}, which replaces the separate per-finger decoders with a single $24\times32$ decoder and also removes panel exchange, convolutional readout, and side embeddings; and
(3) \underline{\smash{\method w/o convolutional readout}}, which replaces the convolutional readout with pointwise MLP heads while retaining the dual-branch decoder and panel exchange.
We evaluate tactile prediction quality for all variants. For downstream evaluation on Eraser Wiping and Triangle Peg Insertion, generated images from each variant replace the tactile input to the same policy trained with real tactile observations.

\noindent \textbf{Frozen Representation Analysis.}
To better understand the role of \method features in downstream manipulation, we compare \method Features and Real Tactile through two prediction tasks: future end-effector translation and current contact intensity. We train linear predictors on pooled generator features and encoded real tactile observations, keeping both representations fixed at their pretrained state without policy fine-tuning. Each task uses 40/10/10 training/validation/test trajectories. For future translation prediction, we combine each representation with two-frame CLIP visual features and proprioception to predict future end-effector positions relative to the current pose. We report L2 error in millimeters, averaged over 15 future steps. For contact intensity prediction, we use each representation from the current frame alone to predict the total unitless contact intensity of each finger sensor. We report MAE averaged over both sensors. Both metrics are averaged across test trajectories.

\begin{table}[tp]
\centering
\setlength{\tabcolsep}{3pt}
\renewcommand{\arraystretch}{1.15}
\begin{tabular*}{\linewidth}{@{\extracolsep{\fill}}lcccc@{}}
\toprule
Method & \makecell[c]{LPIPS\\$\downarrow$} &
\makecell[c]{Energy\\Ratio $\to 1$} & \makecell[c]{Contact\\Acc. $\uparrow$} &
\makecell[c]{Panel\\Asym. $\downarrow$} \\
\midrule
Nearest Neighbor & 0.342 & 3.536 & 0.414 & 1.888 \\
MAE & 0.199 & 4.597 & 0.746 & \textbf{0.431} \\
ViTacGen & 0.243 & 2.441 & 0.659 & 1.348 \\
\method & \textbf{0.191} & \textbf{1.123} & \textbf{0.816} & 0.738 \\
\midrule
\method w/o panel exchange & 0.192 & 0.661 & 0.774 & 1.175 \\
\method single decoder & 0.216 & 0.122 & 0.616 & 2.225 \\
\method w/o conv. readout & 0.213 & 0.259 & 0.665 & 1.225 \\
\bottomrule
\end{tabular*}
\caption[Tactile prediction quality on the xArm test set]{\textbf{Tactile prediction quality on the xArm test set.} The bottom three rows show \method ablations. Bold indicates the best value per metric.}
\label{tab:prediction_combined}
\end{table}

\section{Results}
\label{sec:results}

\begin{figure*}[t]
\centering
\includegraphics[width=\linewidth]{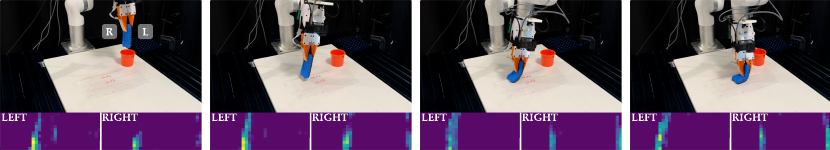}
\caption{
  \textbf{Eraser Wiping rollout with \method Images.} From left to right, the images show the robot bringing the eraser into contact with the table and wiping. The L/R markers on the gripper correspond to the tactile images labeled LEFT/RIGHT below each image. These generated images show changes in contact location and intensity during the rollout.
}
\label{fig:rollout}
\end{figure*}

\begin{table*}[tp]
\centering
\begin{tabular*}{\linewidth}{@{\hskip 4pt\extracolsep{\fill}}llcccc}
\toprule
Method & Train $\rightarrow$ Deploy &
\makecell[c]{Tube\\Insertion} & \makecell[c]{Cup\\Nesting} &
\makecell[c]{Eraser\\Wiping} & \makecell[c]{Triangle Peg\\Insertion} \\
\midrule
Vision Only & None $\rightarrow$ None & 35.0\% & 27.5\% & 57.5\% & 60.0\% \\
DINOv2 Features & DINO $\rightarrow$ DINO & 37.5\% & 22.5\% & 60.0\% & 55.0\% \\
Real Tactile & Real $\rightarrow$ Real & 52.5\% & \textbf{42.5\%} & \textbf{80.0\%} & 77.5\% \\
\method Images & Real $\rightarrow$ Predicted & 45.0\% & 40.0\% & 72.5\% & 62.5\% \\
\method Images & Predicted $\rightarrow$ Predicted & 40.0\% & \textbf{42.5\%} & 75.0\% & 67.5\% \\
\method Features & Latent $\rightarrow$ Latent & \textbf{55.0\%} & \textbf{42.5\%} & 77.5\% & \textbf{82.5\%} \\
\bottomrule
\end{tabular*}
\caption[Final task success rates]{\textbf{Final task success rates.} Each policy configuration is evaluated over 40 trials per task. All policies receive RGB and proprioception; Train $\rightarrow$ Deploy specifies the additional input used during policy training and deployment. Bold indicates the highest success rate per task, including ties.}
\label{tab:policy_main}
\end{table*}

\subsection{Tactile Prediction Quality}
\label{ssec:tactile_quality}

Table~\ref{tab:prediction_combined} reports tactile prediction results on the xArm test set, with qualitative examples in Figure~\ref{fig:xarm_ablation}. \method achieves lower LPIPS, an Energy Ratio closer to 1, and higher Contact Accuracy than Nearest Neighbor, MAE, and the ViTacGen baseline. Compared with ViTacGen, \method reduces LPIPS from 0.243 to 0.191 and improves Contact Accuracy from 0.659 to 0.816, while its Energy Ratio of 1.123 indicates closer agreement with ground-truth total contact intensity than ViTacGen's 2.441. \method also achieves a lower Panel Asymmetry Error than ViTacGen. MAE achieves a lower Panel Asymmetry Error than \method, but its Energy Ratio of 4.597 indicates substantial total-intensity overestimation. These results highlight the complementary roles of contact detection, total intensity, and relative left--right intensity in evaluating tactile prediction quality.

The ablations in Table~\ref{tab:prediction_combined} (bottom) and Figure~\ref{fig:xarm_ablation} reveal prediction errors when architectural components are removed. Removing panel exchange leads to errors in relative left--right contact intensity and misses asymmetric contact patterns, with Panel Asymmetry Error increasing from 0.738 to 1.175.
The single-decoder variant severely underestimates tactile magnitude, with Contact Accuracy falling to 0.616 and Panel Asymmetry Error increasing to 2.225; this variant also removes panel exchange, convolutional readout, and side embeddings.
Removing the convolutional readout reduces Contact Accuracy to 0.665, suggesting that local spatial processing improves contact prediction compared with pointwise decoding.
To assess sensitivity to contact thresholds, we compare \method with all three ablated variants using minimum active-cell counts of $\{6, 8, 10, 12, 14, 16\}$ and intensity thresholds of $\{0.10, 0.15, 0.20, 0.25, 0.30\}$. \method achieves higher Contact Accuracy across all 30 combinations.

\subsection{Downstream Policy Success Rate}
\label{ssec:policy_success}

Table~\ref{tab:policy_main} summarizes final task success for the six configurations. Across all four tasks, Real Tactile improves over Vision Only. \method Images achieves higher observed success rates than Vision Only across all four tasks under both the Real$\rightarrow$Predicted and Predicted$\rightarrow$Predicted configurations. \method Features achieves higher observed success rates than both Vision Only and DINOv2 Features across all four tasks, with results close to Real Tactile: higher on Tube Insertion and Triangle Peg Insertion, equal on Cup Nesting, and slightly lower on Eraser Wiping.
Figure~\ref{fig:rollout} illustrates an Eraser Wiping rollout with \method Images and the corresponding generated tactile images.

Both the Predicted$\rightarrow$Predicted image configuration and \method Features support downstream policy training and deployment without real tactile observations. The feature interface directly supplies intermediate generator features to the policy, bypassing explicit tactile-image prediction and re-encoding. Both configurations still rely on paired RGB--tactile data for generator pretraining. The approximately 20\,ms generator latency on a single NVIDIA RTX 4090 supports real-time closed-loop control.

\begin{table}[tp]
\centering
\renewcommand{\arraystretch}{1.15}
\begin{tabular*}{\linewidth}{@{\extracolsep{\fill}}lcc@{}}
\toprule
Generator variant & \makecell[c]{Eraser\\Wiping} & \makecell[c]{Triangle Peg\\Insertion} \\
\midrule
\method & \textbf{75\%} & \textbf{70\%} \\
\method w/o panel exchange & \textbf{75\%} & 55\% \\
\method single decoder & 70\% & 55\% \\
\method w/o conv. readout & \textbf{75\%} & 60\% \\
\bottomrule
\end{tabular*}
\caption[Policy success with generator ablations]{\textbf{Policy success with generator ablations.} Full \method and its ablated variants are each evaluated over 20 trials per task. The full-model trials are the first 20 of the 40 Real$\rightarrow$Predicted trials in Table~\ref{tab:policy_main}. Generator variants share the same real-tactile-trained policy per task. Bold indicates the highest success rate per task, including ties.}
\label{tab:ablation_rollout}
\end{table}

Table~\ref{tab:ablation_rollout} evaluates downstream deployment with tactile images generated by ablated versions of \method, using the same real-tactile-trained policy over 20 trials per task. On Triangle Peg Insertion, all ablations reduce final insertion success relative to the full model, with the no-panel-exchange and single-decoder variants showing the largest drop. On Eraser Wiping, the no-panel-exchange and no-convolutional-readout variants match the full model's 75\% success rate, while the single-decoder variant achieves 70\%. In our robot trials, the single-decoder variant often produces slower, less stable insertion attempts with repeated retries, consistent with its poor tactile prediction quality in Table~\ref{tab:prediction_combined}.

\subsection{Effect of Pretraining Data Amount}

We train the generator on random 10/25/50\% subsets sampled by demonstration, with a matched optimization budget and identical checkpoint selection (Table~\ref{tab:pretraining_scaling}). Using half the data matches the full model's LPIPS and nearly matches its Contact Accuracy, suggesting diminishing returns on these metrics beyond 50\%. Full data gives the best Panel Asymmetry Error, while half the data gives the Energy Ratio closest to 1.

\begin{table}[tp]
\centering
\begin{tabular*}{\linewidth}{@{\extracolsep{\fill}}lcccc@{}}
\toprule
\makecell[c]{Paired\\data} & \makecell[c]{LPIPS\\$\downarrow$} &
\makecell[c]{Energy\\Ratio $\to 1$} & \makecell[c]{Contact\\Acc. $\uparrow$} & \makecell[c]{Panel\\Asym. $\downarrow$} \\
\midrule
10\% & 0.208 & 0.763 & 0.726 & 1.001 \\
25\% & 0.192 & 0.623 & 0.790 & 0.860 \\
50\% & \textbf{0.191} & \textbf{0.925} & 0.810 & 0.918 \\
100\% & \textbf{0.191} & 1.123 & \textbf{0.816} & \textbf{0.738} \\
\bottomrule
\end{tabular*}
\caption[Effect of pretraining data amount]{\textbf{Effect of pretraining data amount.} Pretraining subsets are sampled at the demonstration level. All models are evaluated on the xArm test set. Bold indicates the best value per metric, including ties.}
\label{tab:pretraining_scaling}
\end{table}


\subsection{Frozen Representation Analysis}
\label{ssec:frozen_representations}

\begin{table}[tbp]
\centering
\setlength{\tabcolsep}{3pt}
\renewcommand{\arraystretch}{1.15}
\begin{tabular*}{\linewidth}{@{\extracolsep{\fill}}lcccc@{}}
\toprule
& \multicolumn{2}{c}{\makecell[c]{Future translation\\L2 error (mm) $\downarrow$}} &
\multicolumn{2}{c}{\makecell[c]{Contact intensity\\MAE $\downarrow$}} \\
\cmidrule(lr){2-3}\cmidrule(lr){4-5}
Task & \makecell[c]{Real\\Tactile} & \makecell[c]{\method\\Features} &
\makecell[c]{Real\\Tactile} & \makecell[c]{\method\\Features} \\
\midrule
Tube Insertion & 18.36 & \textbf{16.76} & \textbf{0.827} & 1.508 \\
Cup Nesting & 33.33 & \textbf{29.74} & \textbf{1.416} & 2.438 \\
Eraser Wiping & 31.68 & \textbf{29.33} & \textbf{1.293} & 2.245 \\
Triangle Peg Insertion & 29.37 & \textbf{24.60} & \textbf{1.334} & 1.750 \\
\bottomrule
\end{tabular*}
\caption[Frozen representation analysis]{\textbf{Frozen representation analysis.} Bold indicates the lower error for each task and metric.}
\label{tab:frozen_probes}
\end{table}

Across all four tasks, \method Features yields lower future translation L2 error than Real Tactile, while Real Tactile yields lower contact intensity MAE (Table~\ref{tab:frozen_probes}). This pattern suggests that \method features provide motion-related cues that may support downstream manipulation despite less accurate contact intensity estimation than with Real Tactile.

\section{Limitations}
\label{sec:limitations}
\vspace{-2pt}

While promising, \method has limitations that motivate future work. First, \method depends on the variety and quality of available paired visuo-tactile data for generator pretraining. Second, our method relies on visual cues in RGB observations. Heavy occlusion can make \method fail to infer tactile information, and contact properties that leave no visual trace cannot be recovered from RGB alone. Additional camera viewpoints may help reduce occlusion. Third, \method predicts each frame independently without explicitly enforcing temporal consistency. During sustained contact, its generated tactile images can fluctuate even when real tactile readings remain nearly static. Finally, we only consider pressure-based tactile sensors~\cite{huang2024_3DViTac,huang2026flexitac}; robustness to sensor wear and transfer across sensor types remain untested. Extending \method to synthesize other types of tactile data such as magnetic~\cite{bhirangi2021reskin,bhirangi2025anyskin} or gel-based~\cite{donlon2018gelslim} sensors is promising.

\section{Conclusion}
\label{sec:conclusion}
\vspace{-2pt}

We present \method, a framework that uses paired visuo-tactile data to learn spatial, per-finger contact representations from RGB observations. We provide image and feature interfaces. %
The image interface enables policies trained with either real or generated tactile observations to use generated images at deployment, while the feature interface supports downstream policy training and deployment without real tactile input.
Across four contact-rich real-robot tasks, both interfaces achieve higher observed success rates than vision-only policies, with \method Features improving success by 15--22.5 percentage points.

\section*{Acknowledgments}
Daniel Seita acknowledges generous support from Starpilot, NSF grant \#2447397, Google DeepMind, NVIDIA, Dexmate, Samsung Research America, and Honda Research Institute. This work was also partially supported by Analog Devices and the Qualcomm Innovation Fellowship. This article solely reflects the opinions and conclusions of its authors and should not be interpreted as necessarily representing the official policies, either expressed or implied, of the sponsors.

\bibliographystyle{IEEEtran}
\bibliography{example_arxiv_v2}

\clearpage
\pagestyle{plain}
\setcounter{section}{0}
\setcounter{subsection}{0}
\setcounter{figure}{0}
\setcounter{table}{0}
\setcounter{equation}{0}
\renewcommand{\thesection}{S\arabic{section}}
\renewcommand{\thesectiondis}{\thesection.}
\renewcommand{\thesubsection}{\thesection.\arabic{subsection}}
\renewcommand{\thesubsectiondis}{\thesubsection.}
\renewcommand{\thefigure}{S\arabic{figure}}
\renewcommand{\thetable}{S\arabic{table}}
\renewcommand{\theequation}{S\arabic{equation}}
\renewcommand{\theHsection}{supp.\arabic{section}}
\renewcommand{\theHsubsection}{supp.\arabic{section}.\arabic{subsection}}
\renewcommand{\theHfigure}{supp.\arabic{figure}}
\renewcommand{\theHtable}{supp.\arabic{table}}
\renewcommand{\theHequation}{supp.\arabic{equation}}
\twocolumn[{%
\begin{center}
{\Large\bfseries Supplementary Material for \method}\\[4pt]
{\large Generating Tactile Signals from Vision for Visuo-Tactile Manipulation}
\end{center}
\vspace{\baselineskip}
}]

\def\ARXIVWITHSUPPLEMENT{}
\ifdefined\ARXIVWITHSUPPLEMENT
\newcommand{\SUPPLEMENTEND}{}
\else
\documentclass{article}

\usepackage{corl_2026}

\renewcommand{\thefigure}{S\arabic{figure}}
\renewcommand{\thetable}{S\arabic{table}}
\renewcommand{\thesection}{S\arabic{section}}
\renewcommand{\thesubsection}{S\arabic{section}.\arabic{subsection}}

\title{
Supplementary Material for \method: Generating Tactile Signals from Vision\\
for Visuo-Tactile Manipulation
}

\author{
Anonymous Author(s)
}

\begin{document}
\maketitle
\newcommand{\SUPPLEMENTEND}{%
    \clearpage
    \bibliography{example}
    \end{document}}
\fi

\section{Data Collection and Synchronization}
\label{sec:supp_data_sync}

\subsection{Collection Hardware and Online Logging}
\label{ssec:supp_collection_pipeline}

\begin{table}[!htbp]
\centering
\small

\begin{tabular*}{\linewidth}{@{\extracolsep{\fill}}llp{0.56\linewidth}@{}}
\toprule
Stream & Rate & Converted representation \\
\midrule
RGB & 60\,Hz & $224 \times 224 \times 3$ RGB image \\
Tactile & 23\,Hz & $12 \times 64$ tactile image \\
Robot state & 60\,Hz & End-effector pose, gripper width \\
\bottomrule
\end{tabular*}
\caption[Data streams collected in the xArm demonstrations]{\textbf{Data streams collected in the xArm demonstrations.} The two tactile panels are concatenated side by side for storage.}
\label{tab:supp_streams}
\end{table}

We collect xArm data with a GELLO teleoperation setup~\cite{wu2024gello}. The operator moves a passive GELLO arm, and its joint configuration is mapped to commands for the xArm7 robot. The xArm7 is equipped with a UMI-style parallel gripper~\cite{chi2024umi}, a wrist-mounted GoPro camera, and two FlexiTac tactile pads following the tactile sensing setup of Zhu et al.~\cite{zhu2025touchwild,huang2026flexitac,huang2024_3DViTac}. One tactile pad is mounted on each finger, and each pad produces a $12 \times 32$ tactile image encoding contact intensity.

Each demonstration starts from a fixed initial robot pose. Before recording, the xArm7 moves to this pose and the operator aligns the GELLO arm to the same configuration. Because the three data streams operate at different rates (Table~\ref{tab:supp_streams}), we align all streams after recording using corrected timestamps as described in Section~\ref{ssec:supp_synchronization}.

\subsection{Timestamp Alignment and Sensor Synchronization}
\label{ssec:supp_synchronization}

Each raw demonstration is converted into a synchronized episode before training or evaluation. We use RGB video frames as the reference timeline. RGB timestamps are corrected by subtracting the measured GoPro--HDMI capture latency of 0.315\,s, calibrated with a QR-code timing procedure.

Robot states are corrected by 1\,ms and resampled to the RGB timestamps. Tactile timestamps are corrected by 20\,ms and matched to RGB frames by nearest-neighbor timestamp matching.

Episodes are excluded if the maximum RGB-to-tactile matching gap exceeds 100\,ms. During conversion, RGB frames are decoded and fixed visual artifacts are masked. Visible ArUco tags are inpainted when present, and frames are center-cropped and resized to $224 \times 224$. The two $12 \times 32$ tactile pads ($I_t^\text{tac} \in \mathbb{R}^{H' \times W' \times 2}$ in the main paper) are concatenated into a $12 \times 64$ left-right tactile image for storage. The final episodes are saved in a Zarr replay buffer with synchronized RGB observations, tactile images, end-effector pose, gripper width, and episode boundary metadata.

\subsection{Generator Training and Test Data}
\label{ssec:supp_generator_data}

\begin{table*}[t]
\centering
\small

\renewcommand{\arraystretch}{1.2}
\begin{tabular*}{\linewidth}{@{\hskip 4pt\extracolsep{\fill}}lcll@{\hskip 4pt}}
\toprule
Dataset & Demonstrations & Frames / pairs & Use \\
\midrule
Zhu et al.~\cite{zhu2025touchwild} & 2,700+ & $\sim$2.6M pairs & Generator train/eval; policy-encoder pretraining \\
xArm test set & 30 & $\sim$72K frames & Generator test \\
Policy demonstrations & 240 & $\sim$520K frames & Policy training \\
\bottomrule
\end{tabular*}
\caption[Datasets and evaluation sets used in the paper]{\textbf{Datasets and evaluation sets used in the paper.}}
\label{tab:supp_datasets}
\end{table*}

Table~\ref{tab:supp_datasets} summarizes the datasets used in our experiments. We train $G_\phi$ on the dataset from Zhu et al.~\cite{zhu2025touchwild} using an 80/20 train/evaluation split by demonstration. We test out-of-distribution tactile prediction on a separate xArm test set collected with the GELLO pipeline; its objects, tasks, and robot embodiment are independent from the generator training data.

\subsection{Policy Demonstrations and Robot Evaluation}
\label{ssec:supp_policy_data}

For downstream policy learning, we collect 60 xArm demonstrations for each task: Tube Insertion, Cup Nesting, Eraser Wiping, and Triangle Peg Insertion. Each policy episode contains RGB observations, tactile images, and robot trajectory fields, aligned via the timestamp procedure in Section~\ref{ssec:supp_synchronization}. Policy action targets are constructed from these trajectories during policy preprocessing. For real-robot evaluation, policies run at 10\,Hz with the same observation preprocessing used in training. Each of the six configurations in Table~\ref{tab:policy_main} is evaluated over 40 trials per task. Generator deployment ablations use 20 trials per task, as specified in Table~\ref{tab:ablation_rollout}. Trials are interleaved and execution order is randomized as described in the main paper.

\section{\method Architecture Details}
\label{sec:supp_architecture}

\subsection{Visual Feature Extraction}
\label{ssec:supp_visual_features}

RGB observations $I_t^\text{vis}$ are center-cropped and resized to $224 \times 224$, converted to $[0,1]$, and normalized with ImageNet statistics. For fisheye wrist-camera inputs, invalid border pixels are removed with a binary validity mask before spatial augmentation. We discard the CLS token and average the 256 spatial patch tokens across DINOv2-B/14 layers 3, 7, and 11 to combine low-level texture with high-level semantic information. The pretrained visual encoder remains frozen throughout training.

\subsection{Panel-Aware Query Decoder}
\label{ssec:supp_query_decoder}

Each panel branch has 384 learnable queries with factorized row and column positional embeddings. Each branch uses two pre-norm transformer decoder layers with eight attention heads and a hidden dimension of 768. Queries perform self-attention followed by cross-attention to the visual tokens in each decoder layer. After each layer, a gated cross-panel exchange module applies bidirectional cross-attention between the two branches; the exchange gates are initialized near zero so the branches start independent and learn to share information during training.

\subsection{Convolutional Readout and Contact-Aware Features}
\label{ssec:supp_readout_features}

Decoded queries are reshaped to $12 \times 32$ feature maps and projected to 256 channels through a $1 \times 1$ convolution. A learned side embedding distinguishes left from right panels. The shared panel-aware convolutional readout applies three residual depthwise-separable convolution blocks. A $1 \times 1$ contact head predicts per-cell contact probabilities $\hat{c}_t$. These probabilities are concatenated with the spatial features and passed to a $1 \times 1$ intensity head to produce the nonnegative unitless contact-intensity map $\widetilde{I_t^\text{tac}}$. The two panel maps are concatenated into a $12 \times 64$ tactile image for storage.

For the feature interface, we use the per-panel spatial features before the final prediction heads. Consistent with Section~\ref{ssec:augment_data}, we denote these features by $(\bz_t^L,\bz_t^R)=G_\phi^\text{feat}(I_t^\text{vis})$, where $\bz_t^L,\bz_t^R\in\mathbb{R}^{12\times32\times256}$.

\subsection{Architectural Ablations}
\label{ssec:supp_ablations}

We evaluate the following architectural variants while keeping the visual encoder, training losses, and training schedule unchanged:
\begin{itemize}[noitemsep,leftmargin=*]
    \item \textbf{\method w/o panel exchange}: Gated cross-panel attention is removed. The two panel branches decode independently with no left-right communication. The separate per-finger decoders and convolutional readout are unchanged.
    \item \textbf{\method single-decoder variant}: The separate per-finger decoders are replaced by a single $24 \times 32$ decoder that predicts the full tactile image jointly. This variant also removes panel exchange, convolutional readout, and side embeddings.
    \item \textbf{\method w/o convolutional readout}: The convolutional readout is replaced with pointwise MLP heads. The separate per-finger decoders and panel exchange are unchanged.
\end{itemize}

\section{Generator Training and Metric Definitions}
\label{sec:supp_losses_metrics}

\subsection{Contact and Intensity Losses}
\label{ssec:supp_losses}

Let $I_{t,q}^\text{tac}$ and $\widetilde{I_{t,q}^\text{tac}}$ denote the ground-truth and predicted unitless contact intensities at cell $q$. The index $q$ spans the canonical $24\times32$ two-panel tactile grid. Binary contact labels are derived as
\begin{equation}
    c_{t,q} = \mathbf{1}\left[I_{t,q}^\text{tac} > \epsilon_c\right],
    \qquad \epsilon_c = 0.01.
\end{equation}
The generator predicts a contact logit $\ell_{t,q}$ and a nonnegative intensity $\widetilde{I_{t,q}^\text{tac}}$ for each cell, with $\hat{c}_{t,q}=\sigma(\ell_{t,q})$ the predicted contact probability.

The contact objective combines focal loss~\cite{lin2017focal}, soft Dice loss~\cite{milletari2016vnet}, and a logit regularizer:
\begin{equation}
    \mathcal{L}_\text{contact}
    =
    \mathcal{L}_\text{focal}
    + 0.5\,\mathcal{L}_\text{dice}
    + 0.01\,\mathbb{E}_{t,q}\left[\ell_{t,q}^{2}\right].
\end{equation}
Focal loss uses $\alpha=0.75$ and $\gamma=2.0$:
\begin{equation}
    \mathcal{L}_\text{focal}
    =
    \mathbb{E}_{t,q}
    \left[
    \alpha_{t,q}(1-p_{t,q})^\gamma
    \operatorname{BCE}(\ell_{t,q}, c_{t,q})
    \right],
\end{equation}
where $p_{t,q}=\hat{c}_{t,q}$ if $c_{t,q}=1$ and $1-\hat{c}_{t,q}$ otherwise, and $\alpha_{t,q}=\alpha$ for contact cells and $1-\alpha$ for background. The soft Dice term is computed per frame:
\begin{equation}
    \mathcal{L}_\text{dice}
    =
    1 -
    \frac{2\sum_q \hat{c}_{t,q} c_{t,q} + 1}
    {\sum_q \hat{c}_{t,q} + \sum_q c_{t,q} + 1}.
\end{equation}

We use a value-weighted Huber term~\cite{huber1964robust}:
\begin{equation}
    \mathcal{L}_\text{Huber}
    =
    \frac{
    \mathbb{E}_{t,q}\left[w_{t,q}\,\operatorname{Huber}(\widetilde{I_{t,q}^\text{tac}}, I_{t,q}^\text{tac})\right]
    }{
    \mathbb{E}_{t,q}\left[w_{t,q}\right]
    }.
\end{equation}
Contact cells receive weight $4.0\,(I_{t,q}^\text{tac}/0.3832)^{0.5}$, a one-cell dilation boundary band receives weight 2.0, and background cells receive weight 0.25.

The intensity objective includes two regularizers. Background suppression (weight 0.05) penalizes intensity outside the dilated contact region:
\begin{equation}
    \mathcal{L}_\text{bg}
    =
    \mathbb{E}_{t,q \in \Omega_\text{bg}}
    \left[\max(\widetilde{I_{t,q}^\text{tac}},0)\right].
\end{equation}
A normalized shape loss (weight 0.05) compares contact-intensity distributions after normalizing by total contact intensity:
\begin{equation}
    \mathcal{L}_\text{shape}
    =
    \left\|
    \frac{\widetilde{I_t^\text{tac}}}{\sum_q \widetilde{I_{t,q}^\text{tac}} + 10^{-6}}
    -
    \frac{I_t^\text{tac}}{\sum_q I_{t,q}^\text{tac} + 10^{-6}}
    \right\|_1.
\end{equation}
The intensity objective in Equation~\ref{eq:loss} is
\begin{equation}
\begin{aligned}
    \mathcal{L}_\text{intensity}
    &= \mathcal{L}_\text{Huber}\\
    &\quad + \lambda_1\mathcal{L}_\text{bg}
    + \lambda_2\mathcal{L}_\text{shape},
\end{aligned}
\end{equation}
where $\lambda_1=\lambda_2=0.05$. The full objective $\mathcal{L}(\phi)$ is the paired-data expectation in Equation~\ref{eq:loss}.

\begin{figure*}[t]
\centering
\includegraphics[width=0.98\textwidth]{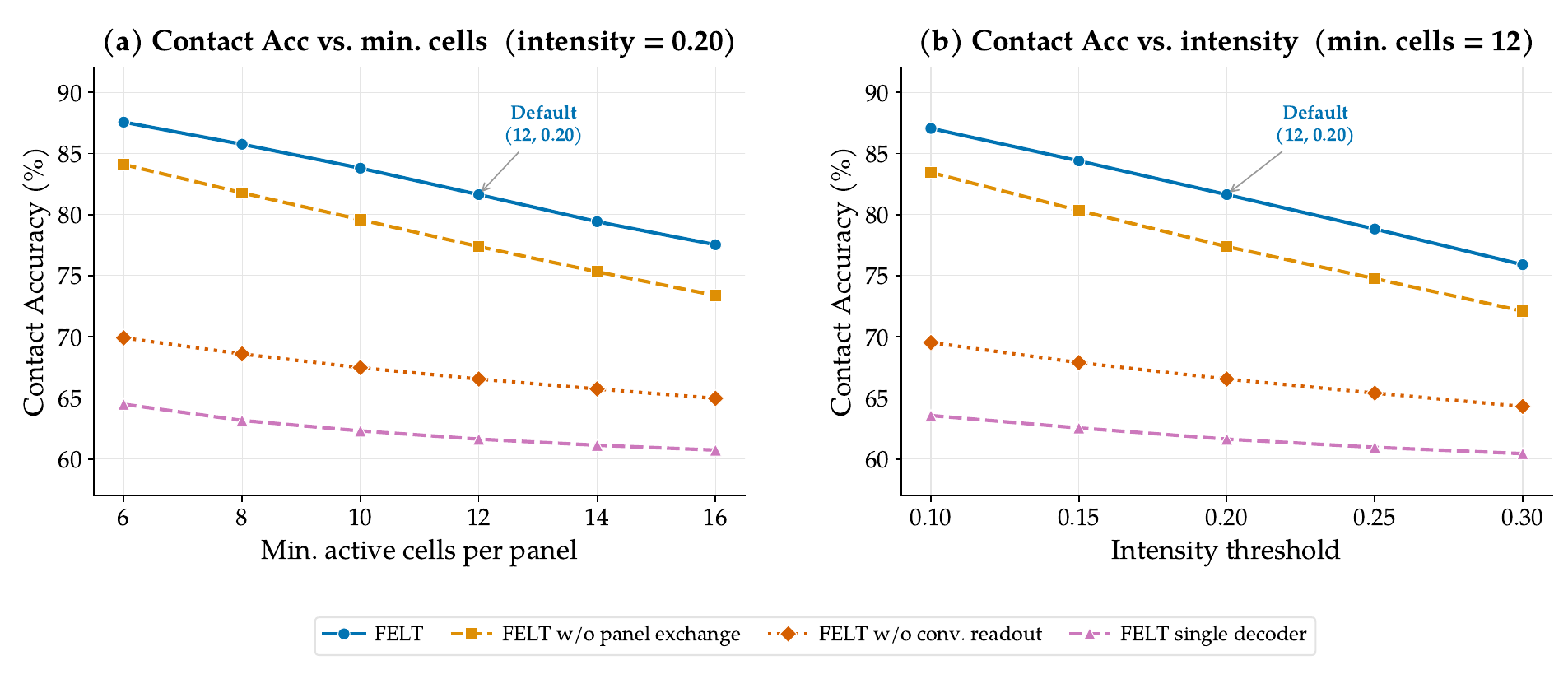}
\caption{
\textbf{Contact Accuracy threshold sensitivity.}
(a) Contact Accuracy as a function of $n_{\text{min}}$, with $\epsilon_\text{eval}$ fixed at 0.20.
(b) Contact Accuracy as a function of $\epsilon_\text{eval}$, with $n_{\text{min}}$ fixed at 12.
The vertical dotted lines mark the default thresholds. The plots show two slices of the 30-combination sweep reported in Section~\ref{ssec:supp_threshold_sensitivity}; \method~achieves higher Contact Accuracy than all three ablated variants across all 30 combinations.
}
\label{fig:supp_frame_acc_sweep}
\end{figure*}

\subsection{Generator Training Protocol}
\label{ssec:supp_generator_training}

\begin{table}[!htbp]
\centering
\small

\begin{tabular*}{\linewidth}{@{\extracolsep{\fill}}p{0.39\linewidth}p{0.56\linewidth}@{}}
\toprule
Setting & Value \\
\midrule
Optimizer & AdamW \\
Learning rate & $2\times 10^{-4}$ \\
Weight decay & 0.003 \\
Batch size & 128 \\
Training epochs & 40 \\
Warmup & 2 epochs (linear) \\
LR schedule & Cosine decay after warmup \\
Precision & AMP (bfloat16) \\
EMA decay & 0.9999 \\
Validation interval & Every 2 epochs \\
Checkpoint selection & Lowest validation tactile-image MAE \\
\bottomrule
\end{tabular*}
\caption[Generator training hyperparameters]{\textbf{Generator training hyperparameters.}}
\label{tab:supp_generator_hparams}
\end{table}

Table~\ref{tab:supp_generator_hparams} lists the generator training settings. The EMA checkpoint is used for tactile prediction evaluation and visualization.

Image augmentations during generator training include color jitter, Gaussian noise (with a standard deviation of 0.03), occasional gamma perturbation, and small in-plane rotations. The fisheye validity mask is applied before spatial augmentation.

\subsection{Tactile Prediction Metrics}
\label{ssec:supp_prediction_metrics}

LPIPS and Contact Accuracy follow the definitions in Section~\ref{ssec:exp_evals}. Below we give explicit formulas for Energy Ratio and Panel Asymmetry Error.

\noindent \textbf{Energy Ratio.}
For each frame with nontrivial ground-truth total contact intensity, we compute
\begin{equation}
    r_t^\text{energy}
    =
    \min\left(\frac{\sum_q \max(\widetilde{I_{t,q}^\text{tac}},0)}
    {\sum_q I_{t,q}^\text{tac} + 10^{-9}}, 10\right),
\end{equation}
and report the mean across valid frames. Values near 1 indicate agreement with ground-truth total contact intensity.

\noindent \textbf{Panel Asymmetry Error.}
For contact frames, let $E_t^L, E_t^R$ and $\hat{E}_t^L, \hat{E}_t^R$ denote the mean contact intensity over ground-truth and predicted left/right panels respectively:
\begin{equation}
    e_t^\text{asym}
    =
    \left|
    \log\frac{\hat{E}_t^L + 10^{-6}}{\hat{E}_t^R + 10^{-6}}
    -
    \log\frac{E_t^L + 10^{-6}}{E_t^R + 10^{-6}}
    \right|.
\end{equation}
The final Panel Asymmetry Error is the mean of $e_t^{\text{asym}}$ across contact frames.

\subsection{Threshold Sensitivity Analysis}
\label{ssec:supp_threshold_sensitivity}

Contact Accuracy (Section~\ref{ssec:supp_prediction_metrics}) depends on two evaluation thresholds: the minimum number of active cells per panel and the intensity threshold for binarizing predicted tactile values. A panel is marked as in contact if at least $n_{\text{min}}$ cells exceed intensity threshold $\epsilon_\text{eval}$. The paper defaults are $n_{\text{min}}=12$ (approximately 3.1\% of the $12\times32$ panel area) and $\epsilon_\text{eval}=0.20$, chosen to suppress isolated noisy activations while remaining sensitive to localized contact.

To verify that our conclusions are not artifacts of these specific thresholds, we sweep $n_{\text{min}} \in \{6, 8, 10, 12, 14, 16\}$ and $\epsilon_\text{eval} \in \{0.10, 0.15, 0.20, 0.25, 0.30\}$, evaluating Contact Accuracy for \method and all three ablations across all $6\times5=30$ parameter combinations, matching the range reported in Section~\ref{ssec:tactile_quality}. Figure~\ref{fig:supp_frame_acc_sweep} shows two slices of this sweep.

\method~achieves higher Contact Accuracy than all three ablated variants across all 30 parameter combinations. These results indicate that its advantage is robust to threshold selection within the evaluated range.

\section{Baseline Implementation Details}
\label{sec:supp_baselines}

All tactile prediction baselines receive only RGB observations at inference and are evaluated against synchronized tactile readings from the xArm test set. At evaluation, ground-truth tactile observations are used only for metric computation.

\subsection{Nearest Neighbor Tactile Retrieval}
\label{ssec:supp_nn}

For each test RGB image, we retrieve the most visually similar training frame and return its paired tactile image. We use the same frozen DINOv2-B/14 encoder and RGB preprocessing as \method (Section~\ref{ssec:supp_visual_features}). The retrieval descriptor is the mean-pooled, $\ell_2$-normalized $16 \times 16$ patch-token grid (256 tokens of dimension 768). At evaluation, we compute cosine similarity between the test descriptor and all training descriptors, retrieve the nearest neighbor, and use its paired tactile image as the prediction. The retrieved tactile array is converted to the canonical two-panel grid before metric computation.

\subsection{Masked Autoencoder Baseline}
\label{ssec:supp_mae}

We use the visuo-tactile MAE architecture from Zhu~et~al.~\cite{zhu2025touchwild} with 100\% tactile masking, so the model receives only learned mask tokens and no real tactile values. The image encoder is a CLIP-pretrained ViT-B/16~\cite{radford2021clip,dosovitskiy2021imageworth16x16words}. The tactile branch encodes the two $12 \times 32$ panels stacked into a $24 \times 32$ grid at the patch level; under full masking, reconstruction relies entirely on RGB context. The model is trained with mean-squared reconstruction error against real tactile images.

We follow the same training configuration from Zhu~et~al.~\cite{zhu2025touchwild}: AdamW with learning rate $3\times 10^{-5}$ for the CLIP encoder and $1\times 10^{-4}$ for remaining modules, weight decay $2\times 10^{-3}$, batch size 64, 5 epochs with linear warmup and cosine decay. We use the EMA checkpoint for evaluation.

\subsection{ViTacGen Baseline}
\label{ssec:supp_vitacgen}

We train ViTacGen~\cite{wu2025vitacgenroboticpushingvisiontotouch} on the same paired visuo-tactile dataset as \method, adjusting its input and output shapes to match our data. We select the checkpoint with the lowest validation tactile-image MAE, using the same checkpoint-selection criterion as \method, and evaluate it on the xArm test set.

\section{Downstream Policy Details}
\label{sec:supp_policy_details}

All downstream experiments use the visuo-tactile Diffusion Policy from Zhu~et~al.~\cite{zhu2025touchwild} with a conditional 1D UNet backbone~\cite{chi2025diffusion}. All configurations share the same RGB encoder, proprioceptive input, UNet backbone, and training protocol; only the tactile pathway varies.

\subsection{Observation Configurations}
\label{ssec:supp_observation_configs}

\noindent \textbf{Shared components.}
The RGB pathway uses a CLIP ViT-B/16 encoder~\cite{radford2021clip,dosovitskiy2021imageworth16x16words} fine-tuned during policy training. A learnable \texttt{AttentionPool2d} aggregates 196 spatial tokens into a 768-dim feature per frame. With an observation horizon of 2 frames, the RGB pathway produces a 1536-dim feature. The proprioceptive pathway concatenates end-effector position (3), 6D rotation (6), gripper width (1), and rotation relative to the start pose (6), yielding 32 dims for the two-frame window. All tactile configurations fuse a 768-dim tactile token with the CLIP CLS token via bidirectional cross-attention, keeping the global condition at 1568 dims.

\begin{table*}[t]
\centering
\small

\begin{tabular*}{\linewidth}{@{\hskip 4pt\extracolsep{\fill}}clllcc@{\hskip 4pt}}
\toprule
& \multirow{2}{*}{Configuration} & \multicolumn{2}{c}{Additional Policy Input} & \multicolumn{2}{c}{Real Sensor Required} \\
\cmidrule(lr){3-4} \cmidrule(lr){5-6}
& & Train & Deploy & Train & Deploy \\
\midrule
(1) & Vision Only & --- & --- & \xmark & \xmark \\
(2) & DINOv2 Features & DINOv2 features & DINOv2 features & \xmark & \xmark \\
(3) & Real Tactile & Real sensor & Real sensor & \cmark & \cmark \\
(4) & \method Images (Real$\rightarrow$Predicted) & Real sensor & $G_\phi$ images & \cmark & \xmark \\
(5) & \method Images (Predicted$\rightarrow$Predicted) & $G_\phi$ images & $G_\phi$ images & \xmark & \xmark \\
(6) & \method Features & $G_\phi^\text{feat}$ latents & $G_\phi^\text{feat}$ latents & \xmark & \xmark \\
\midrule
(7) & Zero Tactile & Real sensor & Zeros & \cmark & \xmark \\
\bottomrule
\end{tabular*}
\caption{\textbf{Observation configurations for downstream policy experiments.} All configurations also receive RGB and proprioception. Inputs and sensor requirements refer to downstream policy training and deployment; paired visuo-tactile data are required for generator pretraining. Configurations (1)--(6) correspond to Table~\ref{tab:policy_main}, and (7) is an additional comparison.}
\label{tab:supp_obs_configs}
\end{table*}

Table~\ref{tab:supp_obs_configs} summarizes all configurations.

\noindent \textbf{(1) Vision Only.}
No tactile pathway. The global condition is 1568 dims from RGB and proprioception.

\noindent \textbf{(2) DINOv2 Features.}
The tactile token is replaced by a mean-pooled, linearly projected frozen DINOv2-B/14 feature (768 dims) entering the same cross-attention fusion.

\noindent \textbf{(3) Real Tactile.}
Follows~\cite{zhu2025touchwild}. The two $12{\times}32$ FlexiTac pads are stacked into a $24{\times}32$ grid, mapped to a viridis color image, and encoded by a SimpleCNN to a 768-dim tactile token that enters cross-attention fusion with the CLS token.

\noindent \textbf{(4) \method Images (Real$\rightarrow$Predicted).}
Trained identically to (3). At deployment, the physical sensor reading is replaced by $\widetilde{I^\text{tac}_t} = G_\phi(I^\text{vis}_t)$, processed through the same SimpleCNN pathway. The pretrained tactile generator remains frozen during downstream policy training and deployment. Generator latency is ${\sim}$20\,ms on an RTX 4090.

\noindent \textbf{(5) \method Images (Predicted$\rightarrow$Predicted).}
Both downstream policy training and deployment use \method-generated tactile images. The pretrained tactile generator remains frozen during both stages.

\noindent \textbf{(6) \method Features.}
No real tactile data are used during downstream policy training or deployment. Per-panel latent maps $(\bz_t^L,\bz_t^R)=G_\phi^\text{feat}(I_t^\text{vis})$, with $\bz_t^L,\bz_t^R\in\mathbb{R}^{12\times32\times256}$, are spatially mean-pooled, concatenated to 512 dims, and linearly projected to a 768-dim token replacing the SimpleCNN output. The DINOv2 encoder is frozen; the tactile decoder is fine-tuned. CLIP processes the same RGB image independently.

\noindent \textbf{(7) Zero Tactile (Section~\ref{sec:supp_deployment_comparison}).}
Trained identically to (3). At deployment, the tactile input is all zeros.

\subsection{Policy Training Hyperparameters}
\label{ssec:supp_policy_hparams}

\begin{table}[!htbp]
\centering
\small

\begin{tabular*}{\linewidth}{@{\extracolsep{\fill}}p{0.54\linewidth}p{0.41\linewidth}@{}}
\toprule
Setting & Value \\
\midrule
Optimizer & AdamW \\
Learning rate (UNet) & $3\times 10^{-4}$ \\
Learning rate (tactile encoder) & $3\times 10^{-5}$ \\
Learning rate (CLIP encoder) & $2\times 10^{-5}$ \\
Betas & (0.95, 0.999) \\
Weight decay & $1\times 10^{-6}$ \\
Batch size & 32 \\
Training epochs & 60 \\
Warmup steps & 2{,}000 (linear) \\
Learning-rate schedule & Cosine decay after warmup \\
Precision & bfloat16 \\
Maximum EMA decay & 0.9999 \\
\midrule
Noise scheduler & DDIM \\
Training diffusion steps & 50 \\
Inference diffusion steps & 16 \\
Beta schedule & squaredcos\_cap\_v2 \\
Prediction type & $\epsilon$ (noise) \\
Input perturbation & 0.1 \\
\midrule
Observation horizon & 2 frames \\
Action chunk length & 16 steps \\
Steps executed per inference & 6 \\
Control frequency & 10\,Hz \\
Observation downsampling steps & 6 \\
\bottomrule
\end{tabular*}
\caption[Policy training hyperparameters]{\textbf{Policy training hyperparameters.}}
\label{tab:supp_policy_hparams}
\end{table}

Table~\ref{tab:supp_policy_hparams} lists all hyperparameters. The diffusion backbone is a ConditionalUnet1D~\cite{chi2025diffusion} with channel widths [256, 512, 1024] and FiLM conditioning. The action vector has 10 dimensions: end-effector position delta (3), 6D rotation (6), and gripper width (1). Position and gripper width are range-normalized to $[-1, 1]$. The policy predicts 16-step action chunks (1.6\,s at 10\,Hz); the first 6 steps are executed before re-querying.

Differential learning rates are used: $3\times 10^{-4}$ for the UNet, $3\times 10^{-5}$ for tactile encoder modules, and $2\times 10^{-5}$ for the CLIP encoder. The EMA model is used for all evaluation and deployment.

\noindent \textbf{Data augmentation.}
During training, RGB images undergo random crop (ratio 0.95, resized to $224{\times}224$) and color jitter (brightness 0.3, contrast 0.4, saturation 0.5, hue 0.08), followed by ImageNet normalization. For branches that encode real or generated tactile images, the tactile images pass through the same pipeline after viridis color mapping. This tactile-image preprocessing does not apply to the \method Features or DINOv2 Features inputs. At deployment, only center crop and normalization are applied to image inputs.

\subsection{Real-Robot Evaluation Protocol}
\label{ssec:supp_robot_eval}

Each of the six configurations in Table~\ref{tab:policy_main} is evaluated over 40 trials per task. Generator deployment ablations use 20 trials per task (Table~\ref{tab:ablation_rollout}). The additional Zero Tactile comparison in Section~\ref{sec:supp_deployment_comparison} uses the original 20-trial-per-task evaluation. Trials are interleaved across configurations in randomized order, re-randomized at each cycle, to reduce bias from workspace drift or operator fatigue~\cite{kressgazit2024robotlearning}. Each trial starts from a fixed robot pose with a randomized object configuration. The maximum rollout duration is 120\,s.

\noindent \textbf{Tube Insertion.}
A thin plastic test tube leans against the edge of a paper box at an angle, with its tail end facing the robot. The box also contains background objects (a pipette-tip box and a thick test tube) that are not manipulated. The box position is fixed to ensure reliable grasping. A separate tube rack is placed at a fixed location on the workspace. The robot grasps the tube vertically; because the tube rests at an angle, it is initially misaligned with the gripper. The robot retracts to the start pose, approaches the tube rack, and uses contact between the open end of the tube and the rack to reorient the tube to a vertical orientation. After reorientation, the robot retracts again, confirms the rack position, rotates the gripper parallel to the table, and inserts the tube downward into the rack. During insertion the rack is heavily occluded by the tube and gripper. \textit{Stages}: grasp, reorientation, insertion. \textit{Success}: the tube remains upright in the rack after release.

\noindent \textbf{Cup Nesting.}
Four transparent plastic cups (bottoms painted red for visibility) are placed upside down on the table, distributed approximately uniformly along the table width. The first cup is placed near the robot in a small fixed range for reliable grasping; cups 2--4 are randomized within 20\,cm in depth (toward/away from the robot). For each nesting step, the robot grasps the current stack, retracts to the start pose, locates the next target cup, rotates the gripper parallel to the table, and lowers the stack onto the target. During nesting, the target cup is heavily occluded by the carried cup and gripper. \textit{Stages}: Pick 1st, Nest $1{\to}2$, Nest $2{\to}3$, Nest $3{\to}4$. \textit{Success}: all four cups fully nested and stable after release.

\newpage
\noindent \textbf{Eraser Wiping.}
A whiteboard eraser sits in a small red cup with a fixed orientation; the cup position is slightly randomized. Three strips of transparent tape are placed parallel on the table surface; the bottom two strips each have the word ``Test'' written in marker, with text position slightly randomized. The text can be erased with moderate downward pressure. The robot grasps the eraser, retracts to the start pose, locates remaining text, presses the eraser onto the tape, and wipes from right to left. After each pass the robot retracts to observe whether text remains and repeats if necessary. A maximum of five wipe passes is allowed. \textit{Stages}: grasp, cumulative erasure at ${\leq}2$, ${\leq}3$, and ${\leq}5$ passes. \textit{Success}: all marker text completely erased within five passes.

\noindent \textbf{Triangle Peg Insertion.}
A triangular peg (3\,cm side length) lies on the table with its orientation randomized over 360$^\circ$ and its position slightly randomized. A matching triangular hole with 2\,mm clearance is fixed in position and orientation on the table. The robot grasps the peg, lifts it, locates the hole, rotates the gripper to align the peg with the hole, and inserts downward. During insertion, the hole is partially occluded by the peg and gripper. \textit{Stages}: grasp, insertion. \textit{Success}: the peg remains stably seated in the hole after release.

\begin{figure*}[t]
\centering
\includegraphics[width=\linewidth]{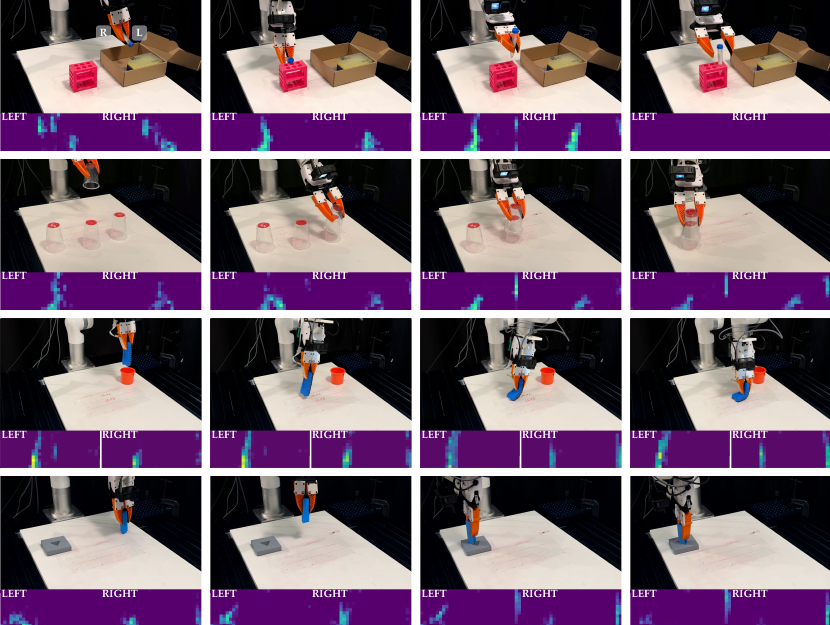}
\caption{
  \textbf{Successful \method Images (Real$\rightarrow$Predicted) rollouts across all four tasks.}
  Each row shows a representative successful trial for one task (top to bottom: Tube Insertion, Cup Nesting, Eraser Wiping, Triangle Peg Insertion), with four time steps progressing left to right. Below each RGB frame, the generated left and right tactile images are displayed.
}
\label{fig:supp_rollout_success}
\end{figure*}

\section{Additional Deployment Comparison}
\label{sec:supp_deployment_comparison}

We additionally evaluate Zero Tactile, which replaces real tactile observations with all-zero input at deployment for a policy trained with real tactile data. This comparison uses the original evaluation with 20 trials per configuration per task, including the corresponding Vision Only and Real Tactile results, rather than the updated 40-trial results in Table~\ref{tab:policy_main}. For Tube Insertion, Cup Nesting, Eraser Wiping, and Triangle Peg Insertion, respectively, Zero Tactile achieves final success rates of 20\%, 20\%, 50\%, and 55\%, compared with 55\%, 35\%, 90\%, and 70\% for Real Tactile, and 40\%, 25\%, 65\%, and 50\% for Vision Only. Zero Tactile has lower observed success rates than Real Tactile on all four tasks and Vision Only on three tasks. These results suggest that the real-tactile-trained policy is sensitive to replacing its tactile input with zeros.

\newpage
\section{Qualitative Rollouts and Failure Cases}
\label{sec:supp_rollouts_failures}

\subsection{Successful Rollouts}
\label{ssec:supp_success_rollouts}

\begin{figure*}[t]
\centering
\includegraphics[width=\linewidth]{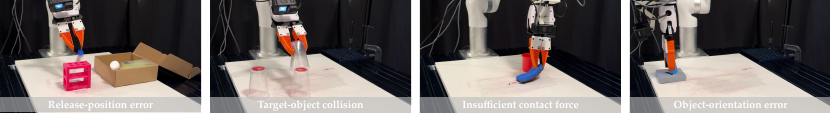}
\caption{
  \textbf{Representative failure cases across all four tasks.}
  From left to right: \emph{Release-position error} (Tube Insertion), where the tube is released outside the rack; \emph{Target-object collision} (Cup Nesting), where the carried cup collides with the target cup and knocks it over; \emph{Insufficient contact force} (Eraser Wiping), where marker text remains after the allowed five passes; and \emph{Object-orientation error} (Triangle Peg Insertion), where the peg is misaligned with the triangular hole.
}
\label{fig:supp_rollout_fail}
\end{figure*}

Figure~\ref{fig:supp_rollout_success} shows one successful trial per task under the \method Images (Real$\rightarrow$Predicted) configuration. In each row, the generated left and right tactile panels track the evolving contact state across task stages.

\noindent \textbf{Tube Insertion} (row~1). The tactile response appears once the gripper closes on the tube and shifts orientation as the tube contacts the rack during reorientation. When the gripper releases the tube, the tactile signal vanishes, consistent with the loss of finger--object contact.

\noindent \textbf{Cup Nesting} (row~2). Initial grasping produces roughly symmetric contact on both fingers.
\newpage
\noindent As the robot lowers the grasped cup onto the target, the tactile maps reorient to reflect the current grasping position and contact geometry.

\noindent \textbf{Eraser Wiping} (row~3). Grasping the eraser from the cup produces localized contact. During wiping, the generated contact-intensity patterns evolve as the eraser moves across the tape strips.

\noindent \textbf{Triangle Peg Insertion} (row~4). The tactile response emerges upon grasping and remains active as the robot aligns and inserts the peg. During insertion, the generated tactile images exhibit distinct patterns as the peg contacts the hole edges, providing contact feedback despite partial occlusion by the peg and gripper.

\newpage
\subsection{Failure Cases}
\label{ssec:supp_failure_cases}

Figure~\ref{fig:supp_rollout_fail} illustrates one representative failure mode per task.

\noindent \textbf{Release-position error} (Tube Insertion). The robot successfully grasps and reorients the tube but releases it at an incorrect position relative to the rack, causing the tube to fall outside the receptacle. During insertion, the rack is heavily occluded by the tube and gripper, making the release point difficult to estimate from vision alone. This case highlights that contact cues about tube orientation do not fully resolve uncertainty in the release position under occlusion.

\noindent \textbf{Target-object collision} (Cup Nesting). While lowering the carried cup onto the target, the gripper contacts the target cup at an offset, knocking it over before nesting can occur. Tactile input can help identify misaligned contact, but in this failure the carried cup occludes the target during the approach, and collision occurs before the policy corrects the alignment.

\newpage
\noindent \textbf{Insufficient contact force} (Eraser Wiping). The robot grasps the eraser and makes contact with the surface, but leaves marker text unerased after the allowed five passes. This behavior is consistent with insufficient sustained downward pressure during wiping. The case highlights the need to maintain sufficient surface contact and pressure throughout each wipe, beyond simply establishing initial contact.

\noindent \textbf{Object-orientation error} (Triangle Peg Insertion). The peg starts in a random orientation on the table, and the robot must infer the correct rotation for insertion. Contact patterns can provide cues about the grasped peg's orientation, but in the shown failure the policy under-rotates the peg, leaving it misaligned with the triangular hole. This case illustrates the difficulty of converting grasp-contact cues into the precise rotation required for insertion.

\SUPPLEMENTEND

\end{document}